\documentclass[conference]{IEEEtran}
\IEEEoverridecommandlockouts
\usepackage{cite}
\usepackage{amsmath,amssymb,amsfonts}
\usepackage{algorithmic}
\usepackage{graphicx}
\usepackage{textcomp}
\usepackage{xcolor}
\usepackage{booktabs}

\def\BibTeX{{\rm B\kern-.05em{\sc i\kern-.025em b}\kern-.08em
    T\kern-.1667em\lower.7ex\hbox{E}\kern-.125emX}}

\newcommand{\Bmat}{{\boldsymbol B}}

\newcommand{\Kmat}[0]{{{\boldsymbol K}}}

\newcommand{\Qmat}[0]{{{\boldsymbol Q}}}

\newcommand{\Tmat}[0]{{{\boldsymbol T}}}

\newcommand{\Vmat}[0]{{{\boldsymbol V}}}
\newcommand{\Wmat}[0]{{{\boldsymbol W}}}

\newcommand{\fv}{\boldsymbol{f}}

\newcommand{\vv}{\boldsymbol{v}}

\newcommand{\xv}{\boldsymbol{x}}
\newcommand{\yv}{\boldsymbol{y}}

\newcommand{\Phimat}{\boldsymbol{\Phi}}

\begin{document}

\title{Unfolding Framework with Complex-Valued Deformable Attention for High-Quality Computer-Generated Hologram Generation}

\author{Haomiao Zhang$^{1,2}$ $\quad$
        Zhangyuan Li$^{1,2}$ $\quad$
        Yanling Piao$^{3}\quad$
        Zhi Li$^{4,5}\quad$ \\
        Xiaodong Wang$^{1,2}\quad$ 
        Miao Cao$^{1,2}\quad$ 
        Xiongfei Su$^{1,2}\quad$ 
        Qiang Song$^{4}\quad$ 
        Xin Yuan$^{2*}$  \\ 
        $^1$Zhejiang University, Hangzhou, Zhejiang 310027, China.\\
        $^2$School of Engineering, Westlake University, Hangzhou, Zhejiang 310030, China.\\
        $^3$Westlake Institute for Optoelectronics, Hangzhou, Zhejiang 311421, China.\\
        $^4$Greater Bay Area Institute for Innovation, Hunan University, Guangzhou 511300, China.\\
        $^5$School of Physics, Sun Yat-sen University, Guangzhou 510275, China.\\
{\tt\small xyuan@westlake.edu.cn}
}

\maketitle

\begin{abstract}
Computer-generated holography (CGH) has gained wide attention with deep learning-based algorithms. However, due to its nonlinear and ill-posed nature, challenges remain in achieving accurate and stable reconstruction.
Specifically, ($i$) the widely used end-to-end networks treat the reconstruction model as a black box, ignoring underlying physical relationships, which reduces interpretability and flexibility. ($ii$) CNN-based CGH algorithms have limited receptive fields, hindering their ability to capture long-range dependencies and global context. 
 ($iii$) Angular spectrum method (ASM)-based models are constrained to finite near-fields. 
In this paper, we propose a Deep Unfolding Network (DUN) that decomposes gradient descent into two modules: an adaptive bandwidth-preserving model (ABPM) and a phase-domain complex-valued denoiser (PCD), providing more flexibility. ABPM allows for wider working distances compared to ASM-based methods. At the same time, PCD leverages its complex-valued deformable self-attention module to capture global features and enhance performance, achieving a PSNR over 35 dB. Experiments on simulated and real data show state-of-the-art results. Code is available at {https://github.com/HannahZhang1926/Complex-Valued-Deformable-Transformer-for-CGH}.

\end{abstract}

\begin{IEEEkeywords}
Computer-generated holography, Convolutional neural network, Transformer, Deep unfolding network
\end{IEEEkeywords}

\section{Introduction}
\label{sec:intro}
Computer-generated holography (CGH) is a prominent display technology, which has attracted wide interest. Traditional CGH algorithms~\cite{gerchberg1972practical,chen2019gradient,candes2015phase}~ suffer from poor performance and long reconstruction time. Deep learning methods, especially end-to-end (E2E) networks, often surpass traditional model-based approaches in both efficiency and effectiveness, significantly enhancing CGH quality while reducing computational complexity~\cite{shui2022diffraction,liu20234k}. However, E2E algorithms lack flexibility. For example, if the wavelength or propagation distance (optical parameters) change, a new network has to be trained. In contrast, deep unfolding networks (DUNs) offer strong interpretability and flexibility, while achieving strong performance in vision tasks~\cite{gregor2010learning}. Additionally, CNN-based algorithms have limited receptive fields, losing detailed information due to challenges in capturing long-range relationships.  In contrast, Transformer models excel at capturing long-range, non-local relationships, improving reconstruction quality by preserving more global and detailed information~\cite{li2023embedding, zhou2023fourmer}. 

\begin{figure}[t]
  \centering
   \includegraphics[width=0.85\linewidth]{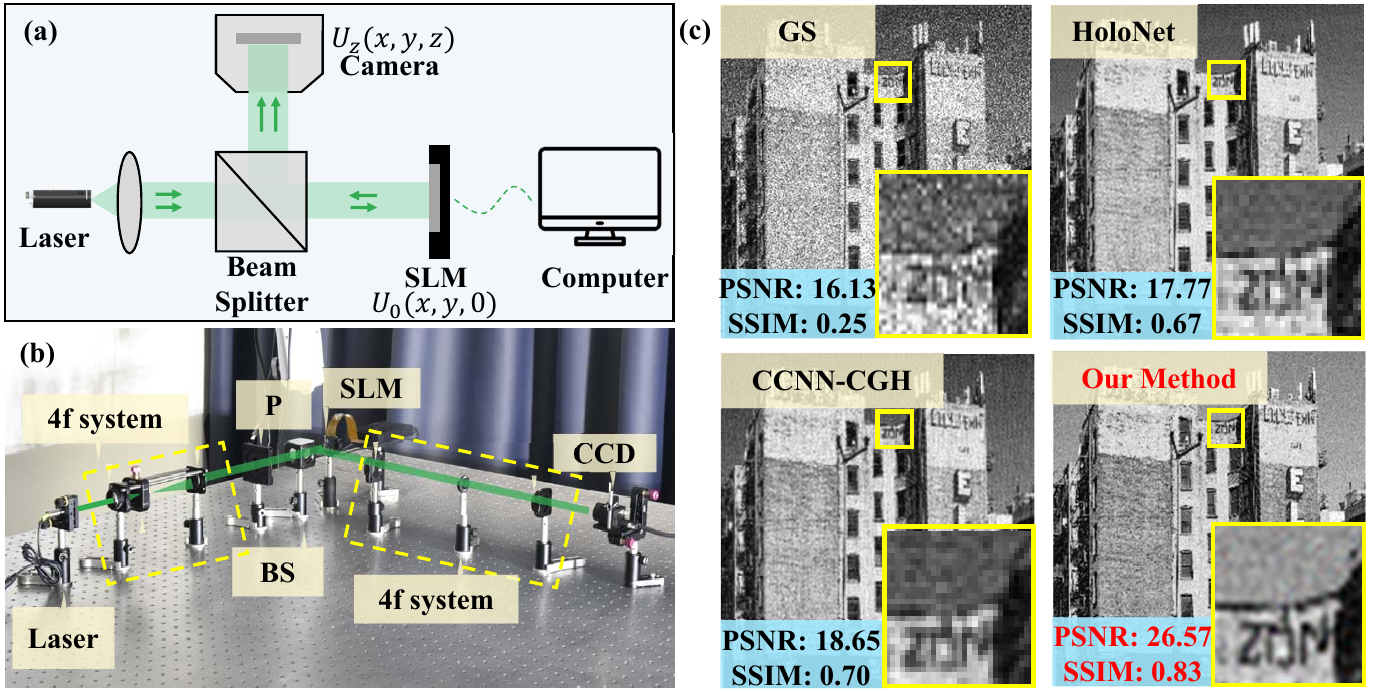}
  \vspace{-3mm}
   \caption{\small {(a) Schematic of CGH model. (b) Experiment setup of our CGH system (BS: beam splitter, CCD: charge-coupled device, P: polarizer). (c) Comparison between the reconstruction results of several representative CGH algorithms in 256×256 resolution.}}
   \label{fig:1}
    \vspace{-5mm}
\end{figure}

\begin{figure*}[t]
  \centering
   \includegraphics[width=0.9\linewidth]{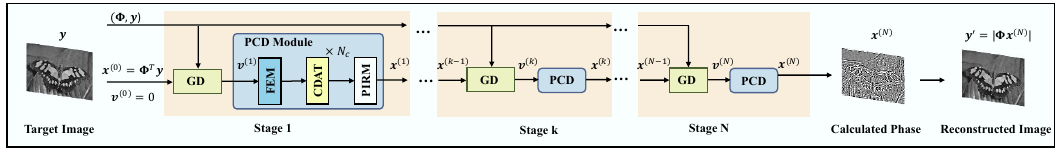}
  \vspace{-3mm}
\caption{\small{The overall framework of our method, containing a deep-unfolding structure with N stages; GD and PCD blocks represent the operation in Eq.\eqref{eqn:gd_ex} and Eq.\eqref{eqn:opt4} respectively.}}
   \label{fig: structure1}
    \vspace{-5mm}
\end{figure*}

Meanwhile, in diffractive optics, angular spectrum method (ASM)-based algorithms are limited by the transfer function~\cite{goodman2005introduction} and require equal-sized object and image planes. Previous research~\cite{matsushima2009band} showed that applying ASM beyond its working distances leads to significant image distortion, such as spectrum aliasing. Efforts have been made to ensure accurate propagation across varying distances, including band-limited angular spectrum method~\cite{matsushima2009band}, non-uniform sampling~\cite{kim2014non}, adaptive sampling~\cite{zhang2020adaptive}, etc. 

In summary, the motivations behind our proposed methods are:
($i$) E2E-CNN algorithms suffer from a lack of flexibility and interpretability and are limited by the respective field~\cite{shui2022diffraction,liu20234k}. 
($ii$) Global attentional mechanisms in existing Complex-valued Transformers have higher memory consumption, making them inefficient for CGH reconstruction~\cite{dong2021signal,eilers2023building,yang2020complex}. 
($iii$) ASM model has a limited propagation distance~\cite{goodman2005introduction}.  
Bearing these concerns in mind, we propose a novel approach for high-quality CGH generation. Our main contributions are summarized as follows.

\vspace{-0.5em}
\begin{itemize}
	\itemsep0em
    \item We propose a DUN framework incorporating global image features and physical constraints in CGH reconstruction, which consists of a gradient descent (GD) module and a deep denoising module to enhance both flexibility and interpretability.
    \item  In the GD module, we replace the widely-used ASM model with an adaptive bandwidth-preserving method (ABPM) based on the sampling theory, enabling a wider working distance.
    \item In the denoising module, we introduce a Phase-domain Complex-valued Denoiser (PCD) with a Complex-valued Deformable Self-Attention (CDSA) mechanism, leveraging spatial self-attention to achieve more efficient and high-quality reconstruction. 
    \item  We built a holographic display system using a phase-only spatial light modulator (SLM) to demonstrate the wide working distance and high performance of our method.
\end{itemize}

\section{Related Work}

\noindent{\bf Model-based Algorithms for CGH:}
Model-based CGH algorithms can be broadly classified into iterative and non-iterative methods. Iterative techniques include the Gerchberg-Saxton (GS) method~\cite{gerchberg1972practical}, Wirtinger holography~\cite{candes2015phase}, and gradient descent~\cite{chen2019gradient}. Non-iterative approaches contain point-based~\cite{tsang2018review}, double phase-amplitude coding~\cite{maimone2017holographic} methods et al. Although these techniques are robust and training-free, their limited performance hinders their application in high-quality holographic displays.

\noindent{\bf Deep Learning Algorithms for CGH:}
Deep learning-based methods for CGH can be categorized into data-driven and physics-driven models. Data-driven CGHs rely on labeled data to learn the mapping between input scenes and output holograms, but they do not fully incorporate the physical model, limiting their performance ~\cite{liu2021deep,shi2021towards,shi2022end,chang2023picture}.  In contrast, model-based E2E-CNN algorithms~\cite{wu2021high,peng2020neural,shui2022diffraction,liu20234k,chang2023complex,zhong2023real} integrate physical diffraction directly into the network, achieving high-fidelity large-scale reconstruction but lacking flexibility and struggling with long-term dependencies, leading to reconstructions below 35 dB. Furthermore, the existing Transformer~\cite{dong2023vision} for CGH reconstruction remains under-explored and suboptimal. In contrast, physics-driven DUNs offer strong performance, clear interpretability, and outperform E2E algorithms in many tasks.~\cite{chen2022physics,qu2024dual}.

\section{Problem Statement}

As depicted in Fig.~\ref{fig:1}(a), the SLM displays a phase-only hologram $U_0(x_0,y_0,0)={\rm exp}({\rm i}\phi(x_0, y_0,0))$. After propagating a distance $z$, the complex amplitude $U_z(x_z,y_z,z)$ is recorded on the camera plane, containing the information from $U_0$. This free propagation process can be represented in the matrix form: $U_z(x_z,y_z,z)=\mathbf{A} U_0(x_0,y_0,0)$, where matrix $\mathbf{A}$ models the discrete propagation from the SLM plane $z=0$ to the camera plane $z=d$. Due to CMOS camera limitations, only the intensity information can be captured. Specifically, the captured digital images are closer to the amplitude $|U_z(x_z, y_z, z)|$ than to the squared intensity $|U_z(x_z, y_z, z)|^2$, as noted in \cite{peng2020neural, gao2023iterative}. Suppose $\yv=U_z(x_z,y_z,z)$ and $\xv=U_z(x_z, y_z, z)$, and the forward model is formulated as $\yv=|\Phimat\xv|$, where $\xv\in \mathbb{C}^{W\times{H}}$, $\yv\in \mathbb{R}^{W\times{H}}$. $\Phimat\in \mathbb{C}^{W\times{H}}$ denotes the propagation matrix, which corresponds to the ABPM model introduced in our article.

\section{Proposed Method}
\subsection{Overall Deep Unfolding Structure}
The general structure of our DUN framework is shown in Fig.\ref{fig: structure1}. To solve the aforementioned ill-posed equation, a regularizer is introduced to constrain the solution space. This regularization term $R(\xv)$ is used to find an estimate $\hat{\xv}$ of $\xv$ by solving the following problem:
\begin{equation}
\hat{\xv}=\textstyle \arg\min_{{\xv}}\frac{1}{2}\bigl\| \yv - \left|\Phimat\xv\right|\bigr\|_{2}^{2} + \lambda R(\xv),
\label{eqn: solve}
\end{equation}
$\lambda$ balances the fidelity term $\frac{1}{2}\bigl\| \yv - \left|\Phimat\xv\right|\bigr\|_{2}^{2}$ and the regularization term. Since the regularization term may be not differentiable or even unknown, it is hard to find a closed-form solution directly.  Alternatively, the Half Quadratic Splitting (HQS) method is adopted, which introduces an auxiliary variable $\vv\in \mathbb{C}^{W\times{H}}$ and turns the unconstrained optimization into an augmented Lagrangian function:
\begin{equation}
(\hat{\xv},\hat{\vv}) =  \textstyle \mathop{\arg\min}\limits_{\xv, \vv}\frac{1}{2}\bigl\| \yv- \left|\Phimat\vv\right|\bigr\|_{2}^{2}+\lambda R(\xv)+\frac{\eta}{2}\bigl\|\vv-\xv\bigr\|^2_2.
\label{eqn: opt2}
\end{equation}
Previous research~\cite{metzler2016denoising} indicated that optimization-based algorithms include : $i$)  Gradient descent and $ii$)  Projection to the signal domain. The first step refines the current estimate by incorporating additional information from the input. The second step constrains the result to the target signal space, typically using a denoiser.  Thus Eq.\eqref{eqn: opt2} can be solved by decoupling $\xv$ and $\vv$ into two iterative sub-problems:
woy\begin{subequations}
\begin{align}
   &\vv^{(k+1)} = \textstyle \mathop{\arg\min}\limits_{\vv} \bigl\| \yv - \left|\Phimat\vv\right|\bigr\|_{2}^{2} + \eta\bigl\|\vv-\xv^{(k)}\bigr\|^2_2, \\
   &\xv^{(k+1)} =  \textstyle \mathop{\arg\min}\limits_{\xv} \frac{1}{2} \bigl\| \xv - \vv^{(k+1)} \bigr\|_2^2 + \lambda R(\xv).
\end{align}\label{eq4}
\end{subequations}

\begin{figure*}[h]
  \centering
   \includegraphics[width=0.85\linewidth]{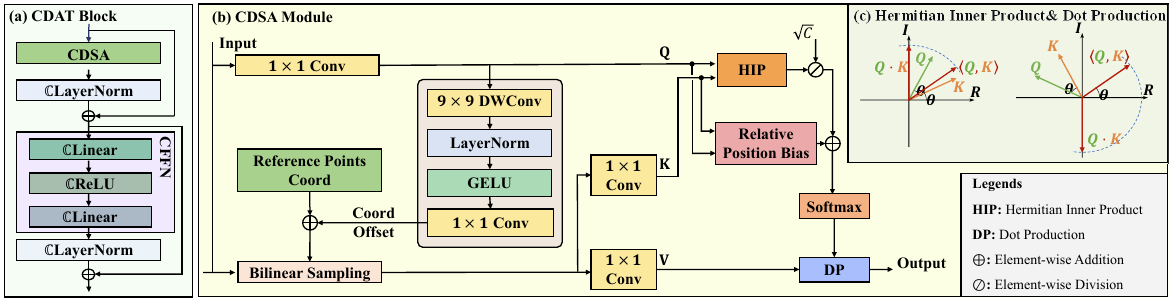}
  \vspace{-3mm}
\caption{\small{(a) The diagram of CDAT block.  (b) Complex-valued Deformable Self-Attention (CDSA) module. (c) An illustration of the difference between Hermitian inner production and dot production. }}
   \label{fig: structure3}
    \vspace{-5mm}
\end{figure*}

For the $\vv$-subproblem, Eq.~(\ref{eq4}a) has a closed-form solution~\cite{gao2023iterative}:
\begin{equation}\label{eqn:gd_ex}
\hspace{-1mm}
\begin{aligned}
\vv^{(k+1)}& = \xv^{(k)} - \rho \nabla_{\xv} F(\xv^{(k)}),\\
&= \textstyle  \xv^{(k)}-\Phimat^H {\rm diag}\left(\frac{{\Phimat\xv}^{(k)}}{|{\Phimat \xv}^{(k)}|}\right)(\yv-|{\Phimat \xv}^{(k)}|),
\end{aligned}
\end{equation}
where $\nabla_{\xv} F(\xv^{(k)})$ denotes the gradient of $F(\xv)=\left|\Phimat \xv\right|$ at the point $\xv = \xv^{(k)}$. $\nabla_{\xv} F(\xv^{(k)})$ is derived as $\nabla_{\xv} F(\xv) = \textstyle \frac{1}{2}\Phimat^H \rm{diag}\left(\frac{\Phimat\xv}{|\Phimat\xv|}\right)(\yv-|\Phimat\xv|)$, 
where $(\cdot)^H$ denotes the conjugate transpose operator. Notice that $\frac{1}{2}$ is absorbed into the constant $\rho$, and $\rho$ is set to $1$. 

For the $\xv$-subproblem, Eq.~(\ref{eq4}b) aims to find $\xv^{(k+1)}$ that is as close as possible to the updated $\vv^{(k+1)}$ while maintaining consistency with previous iteration. Since modeling a proper prior for complex amplitude fields is nontrivial, we implement the mapping from $\vv^{(k+1)}$ to $\xv^{(k+1)}$ using a deep neural network. This module is referred to as the Phase-domain Complex-valued Denoiser (PCD):
\begin{equation}
\xv^{(k+1)}={\rm PCD}(\vv^{(k+1)}).
\label{eqn:opt4}
\end{equation}

Eq.~\eqref{eqn:gd_ex} and Eq.~\eqref{eqn:opt4}  form one iteration (or stage) to solve Eq.\eqref{eqn: solve}. In our DUN framework, this iteration is applied a few times, allowing explicit unfolding and independent network parameters for PCDs at each stage. Learnable PCDs, derived from real data, are more efficient and accurate than traditional hand-crafted priors. By combining the optimization algorithm with the denoising network, our method is both effective and interpretable. Next, we provide a detailed explanation of the propagation matrix $\Phimat$ and $\rm PCD$ blocks. 

\subsection{Adaptive Bandwidth-Preserving Method}\label{Sec:propagation}
To ensure diffraction calculations are performed accurately at all working distances, we rewrite the ASM propagation model as follows:
\begin{equation}\label{eqn: final_prop}
\hspace{-0.1mm}
        \begin{aligned}
	&U_z(x_z,y_z,z)= \\
        &\left\{
	\begin{aligned}
		&\mathcal{F}^{-1} \left\{H(f_x,f_y,z) \mathcal{F} \left\{U_0(x_0,y_0,0) \right\} \right\}, z \leq z_1.\\
		&\mathcal{F}^{-1} \left\{ \mathcal{F} \left\{ h(x_z,y_z,z) \right\} \mathcal{F} \left\{ U_0(x_0,y_0,0) \right\} \right\}, z_1<z \leq  z_2.\\
  &\mathcal{F}^{-1} \left\{ \mathcal{F} \left\{ h'(x_z,y_z,z) \right\} \mathcal{F} \left\{ U_0(x_0,y_0,0) \right\} \right\}, z > z_2.
	\end{aligned}
	\right.
        \end{aligned}
\end{equation}
where $z_1 = {N\delta x / \lambda \sqrt{(\delta x)^2-(\lambda /2)^2}}$ is the ASM threshold, $z_2 = (25 N^4(\delta x)^4 / \lambda)^{\frac{1}{3}}$ is the far field threshold, and $h(x_z,y_z,z)$ and $h'(x_z,y_z,z)$ are the impulse response at various propagation distance. We name this propagation model the adaptive bandwidth-preserving method (ABPM). In far-field propagation, common models such as Fresnel or Fraunhofer increase the image plane size, requiring target image pre-scaling, which reduces fidelity. The band-limited ASM extends propagation slightly further but may lose high-frequency information. The two-step Fresnel model, suitable only for long distances, may lack accuracy. Our ABPM model ensures rigorous sampling across general cases, with the image plane's pixel size matching that of the SLM, avoiding scaling issues. Details are provided in the Supplementary Material (SM).

\subsection{Phase-domain Complex-valued Denoiser} \label{PCD}

As shown in Fig.~\ref{fig: structure1}, the PCD module consists of three parts: $i$)  Feature Extraction Module (FEM), $ii$)  Complex-valued Deformable Attention Transformer (CDAT) blocks and $iii$) Phase Image Recovery Module (PIRM). The FEM and PIRM are explained in detail in the SM.

In Fig.~\ref{fig: structure3}(a), the CDAT module consists of a Complex-valued Deformable Self-Attention (CDSA) module and a Complex-valued Feed-Forward Network (CFFN), both using complex-valued layer normalization for stable training and residual shortcuts for improved efficiency. The CDSA module will be discussed next. The CFFN enhances CDAT's non-linearity and representation ability, featuring two complex-valued linear layers (please refer to the SM) with ReLU activation applied separately to the real and imaginary parts of the input. 

\vspace{-2mm}
\begin{figure*}[h]
  \centering
   \includegraphics[width=0.85\linewidth]{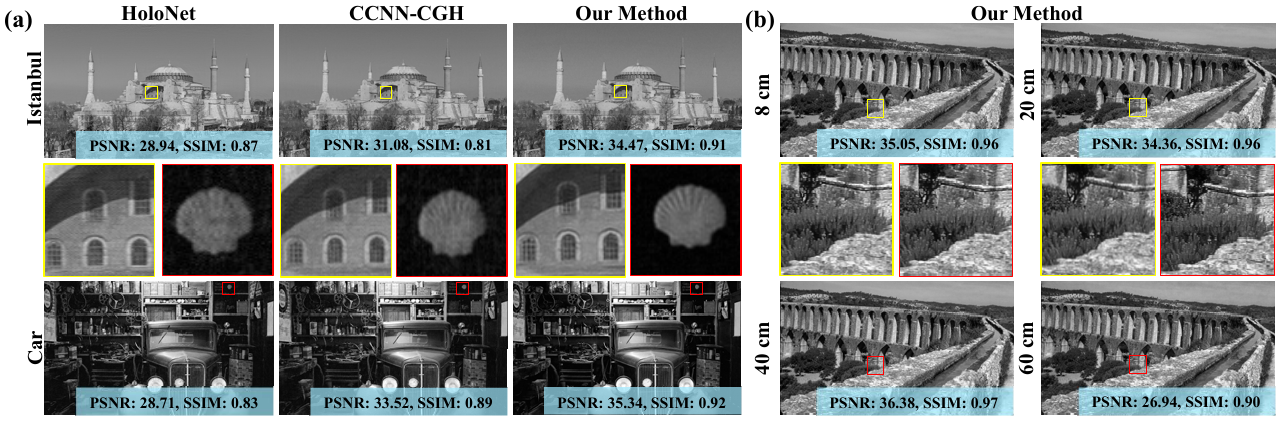}
  \vspace{-4mm}
   \caption{{\small{\bf Simulation}:(a) Numerical reconstructions of CGH generated by different algorithms on four benchmark datasets at 1920\texttimes1080 resolution. (b) CGH reconstruction results of our algorithm at 8 cm, 20 cm, 40 cm, and 60 cm. Zoom in for a better view.}}
   \label{fig:simulation1}
   \vspace{-5mm}
\end{figure*}

\subsection{Complex-valued Deformable Self-Attention module} 
\label{csda}

Previous works~\cite{yang2020complex,eilers2023building} adopt the global scaled dot-product self-attention mechanism, which suffers from high time complexity, making it impractical for large-scale high-resolution CGH generation. The computational bottleneck occurs mainly at the softmax function, thus reducing the input size of the softmax function, i.e. the size of the attention matrix $\Qmat\Kmat^T$, is crucial. In contrast, we employ a local, deformable self-attention mechanism~\cite{xia2022vision} that adaptively selects the attention regions, which alleviates the computational burden concerning image size. 


The structure of our proposed CDSA module is shown in Fig. \ref{fig: structure3}(b). The CDSA module receives the feature map $\fv\in \mathbb{C}^{H/4 \times W/4 \times C}$ from FEM (the FEM will down-sample the size of the input feature map to a quarter). For simplicity, we denote  $\tilde{H}=H/4$ and $\tilde{W}=W/4$. To shrink the size of the attention matrix, the deformable attention mechanism chooses to down-sample $\fv$ before calculating $\Kmat$ and $\Vmat$. The down-sampled points' coordinates are the sum of a fixed reference point and a Query-dependent offset. The reference points are evenly distributed across the original feature map, and their density determines the down-sample rate. In our case, the rate is 64 (8 on each side), reducing the attention matrix by a factor of 64. The offset is generated from $\Qmat$. As shown in Fig.~\ref{fig: structure3}(b), a $1\times 1$ convolution layer converts $\fv$ into the $\Qmat\in\mathbb{C}^{N_Q\times C}$ ($N_Q=\tilde{H}\tilde{W}$), and then a sub-network transforms the $\Qmat$ into coordinate offsets. To be specific, the sub-network consists of a depth-wise convolution layer, a layer normalization, a GELU activation function, and a $1\times 1$ regular convolution layer. The kernel size, stride, and padding size of depth-wise convolution are set to $9\times 9$, $8$, and $4$, which is consistent with the down-sample rate setting. After the last $1\times 1$ convolution, the channel dimension of the feature map is reduced to 2, meaning offset value on the $x$ and $y$ axes respectively. Note that the bias in the $1\times 1$ convolution is omitted to prevent uniform shifts across locations. With the coordinates of the reference points and their offsets, a down-sampled feature map can be interpolated from the original input $\fv$. After this, two $1\times 1$ convolution layers convert this feature map into the $\Kmat\in\mathbb{C}^{N_K\times C}$ and $\Vmat\in\mathbb{C}^{N_V\times C}$ ($N_K=N_V=\tilde{H}\tilde{W}/64$) separately.

When calculating the similarity between $\Qmat$ and $\Kmat$, the dot product operation $\Qmat\Kmat^T$ fails to capture the phase similarity of the complex-valued query-key pairs. To address this, we use the Hermitian inner product to measure their similarity. Given two complex vectors $\xv, \yv\in \mathbb{C}^n$, the Hermitian inner product is defined as $\left< \xv, \yv \right> = {\textstyle\sum^n_{j=1}\xv_j\overline{\yv_j} } ={\textstyle\sum^n_{j=1}|\xv_j||\yv_j|{\rm exp}[{i(\phi_{\xv_j}-\phi_{\yv_j})}]}$,
while the dot product is $\xv\yv^T = {\textstyle\sum^n_{j=1}\xv_j\yv_j } ={\textstyle\sum^n_{j=1}|\xv_j||\yv_j|{\rm exp}[{i(\phi_{\xv_j}+\phi_{\yv_j})}]}$. Although the amplitudes of $\left< \xv, \yv \right>$ and $\xv\yv^T$ are identical, the phase angle of $\left< \xv, \yv \right>$ can reflect the phase discrepancy between $\xv$ and $\yv$, while $\xv\yv^T$ cannot. This proves the rotation invariance of the Hermitian inner product. As depicted in Fig.~\ref{fig: structure3}(c), given two complex inputs $\xv$ and $\yv$ with fixed phase angle discrepancy, the results of the dot product will change when we rotate $\xv$ and $\yv$ synchronously while the results of Hermitian inner product remain the same. 

The Hermitian inner product similarity $\left<\Qmat, \Kmat\right>\in \mathbb{C}^{N_Q\times N_K}$ is scaled by $\sqrt{C}$ and combined with a relative position bias matrix to form a complex attention matrix. During softmax normalization, only the real part is used. Since the similarity dominates the attention matrix (due to initially small values in $\Tmat$), the real part of Hermitian inner product similarity, i.e., $\sum^n_{j=1}|\xv_j||\yv_j|{\rm cos}(\phi_{\xv_j}-\phi_{\yv_j})$, maximizes when ${\rm cos}(\phi_{\xv_j}-\phi_{\yv_j})=1$, equaling to $\phi_{\xv_j}=\phi_{\yv_j}$. It strictly decreases while $|\phi_{\xv_j}-\phi_{\yv_j}|$ grows from 0 to $\pi$ and minimizes at $\phi_{\xv_j}=-\phi_{\yv_j}$. Thus the real part of the complex attention matrix effectively captures the correlation between $Q$ and $K$. 

The overall CDSA mechanism can be written as:
\begin{equation} \label{eqn:CDSA}
\hspace{-0.1mm}
\begin{aligned}
&{\rm CDSA}(\fv)= \\
&\sigma\left(\mathcal{R}\left(\frac{\left< \Wmat_Q\fv, \Wmat_K \phi(\fv)\right>}{\sqrt{C}} + \Bmat\right)\right)\Wmat_V \phi(\fv),
\end{aligned}
\end{equation}
where $\Wmat_Q$, $\Wmat_K$, and $\Wmat_V$ refer to the $1\times 1$ convolution layers and $\phi(\cdot)$ is the deformable down-sample function, $\sigma(\cdot)$ denotes the softmax function, $\mathcal{R}$ refers to extracting the real part. In conclusion, the per-layer complexity of the CDSA mechanism is $O(N_Q N_K C+N_Q N_V C)$, which is much lower than the complexity $O(2N_Q^2 C)$ of scaled dot production self-attention since $N_K, N_V\ll N_Q$ in our setting.

\section{Experimental Results}
\subsection{Implementation Details}\label{Sec:implementation}
Our model is trained on the DIV2K super-resolution dataset~\cite{div2k}, which contains 800 images, and evaluated on 100 randomly selected images from the DIV2K and Flickr2K~\cite{Lim_2017_CVPR_Workshops} datasets. Adam optimizer is used with a learning rate of 1e-4. The model is implemented in PyTorch 2.0 and Python 3.9, trained for 30 epochs on an NVIDIA RTX 8000 GPU. Optical parameters are set to a pixel pitch of 8 $\mu$m, wavelength of 520 nm, and a maximum resolution of 1920$\times$1080.

\vspace{-1mm}
\begin{figure*}[h]
  \centering
   \includegraphics[width=0.85\linewidth]{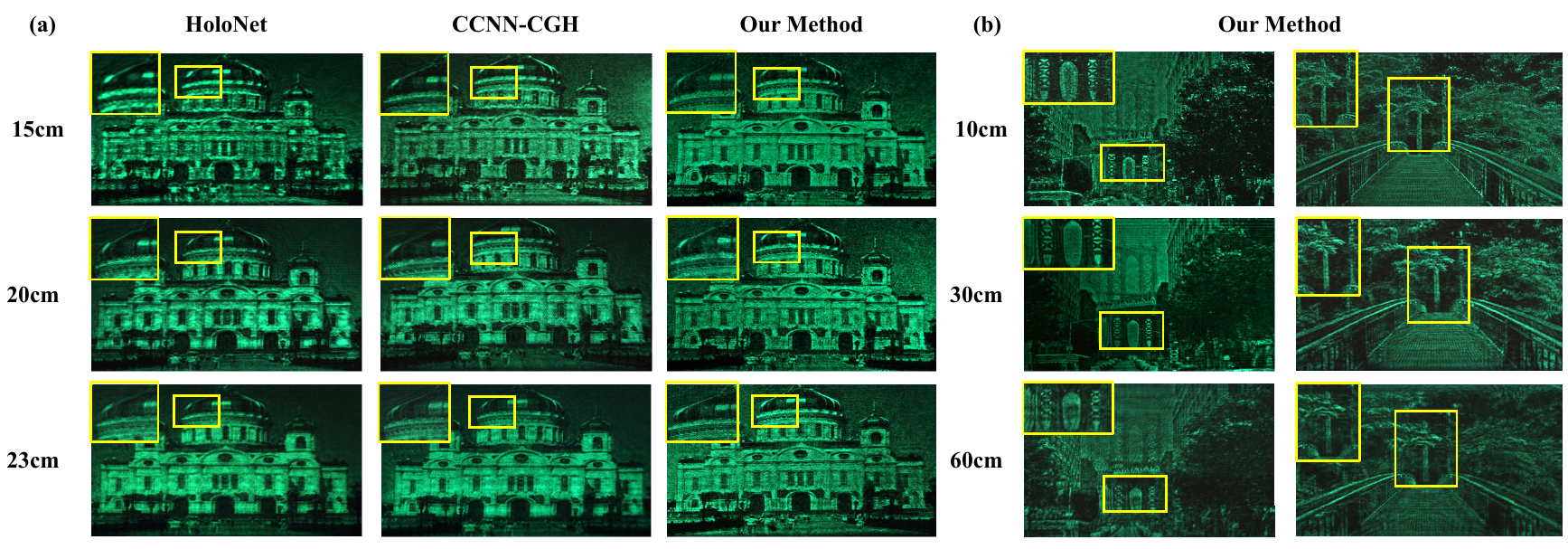}
   \vspace{-3mm}
   \caption{{\small{\bf Real data results}: Reconstructed results of various algorithms at different propagation distances with ASM threshold. (b) Reconstructed results of our algorithm beyond the ASM threshold. Zoom in for a better view.
}}
   \label{fig:experiment1}
\end{figure*}
\vspace{-1mm}

\subsection{Simulation Results \label{Sec:sim}}
Our method is superior to previous CGH methods in terms of both performance and interpretability by integrating a physics-based DUN structure and achieves a better balance of accuracy and flexibility. To verify the performance of our model, we compare our method with the traditional GS algorithm~\cite{gerchberg1972practical}, HoloNet~\cite{peng2020neural} based on CNN with real values, and CCNN-CGH~\cite{zhong2023real} based on CNN with complex values. All models are trained under identical conditions, including the same database, epoch number, and optical setup. The main focus of the comparison is to assess the quality of real-valued images generated from the recovered phase distribution. Peak-Signal-to-Noise-Ratio (PSNR) and Structured Similarity Index Metrics (SSIM)~\cite{wang2004image} are used to evaluate the results.

We compare the performance of various algorithms in the near field. According to Eq.~\eqref{eqn: final_prop}, the threshold of ASM is 23.62 cm. Therefore, we set the diffraction distance at 20 cm. Tab.~\ref{tab:PSNR} shows a comparison of models trained with various methods at a resolution of 1920$\times$1080. The average PSNR and SSIM values were calculated from 100 validation samples using numerical reconstruction. Our method outperforms others, achieving a PSNR improvement of 4.44-15.19 dB and an SSIM increase of 0.03 to 0.4. Moreover, it requires fewer parameters and lower computational complexity (FLOPs), demonstrating superior performance in CGH tasks. Fig.~\ref{fig:simulation1}(a) shows the simulated results of two chosen benchmark images from the test data set ({\tt Istanbul} and {\tt Car}). The results of the GS algorithm are excluded due to its unsatisfactory performance. Our method produces reconstruction results closer to the ground truth, recovering more detail and exhibiting less noise compared with previous algorithms.

\begin{table}[h!]
  \caption{\small{Performance at 1920\texttimes1080 resolution CGH generation.
  }}
  \vspace{-2mm}
  \centering
  \resizebox{.48\textwidth}{!}
{\centering
  \begin{tabular}{cccccc}
    \toprule
     \hline
    Method  & Params & FLOPs(G) &Time(ms)& Avg.PSNR & Avg.SSIM 
\\
 \hline
    \midrule
     GS  
& 
-&- &10640 & 21.26& 0.55
\\
     HoloNet 
& 
2,868,754&328.53  &60& 29.69& 0.90
\\
     CCNN-CGH 
& 42,260&7.21 &16& 32.01& 0.92
\\
     \textbf{Our Method}  & \textbf{19,430}&\textbf{2.63} &140& \textbf{36.45}& \textbf{0.95}\\
      \hline
    \bottomrule
  \end{tabular}}
  \label{tab:PSNR}
\end{table}
\vspace{-1mm}

We also performed a comprehensive evaluation of our proposed ABPM method at different propagation distances from 8 to 60 cm in Fig.\ref{fig:simulation1}(b). The results demonstrate the high quality of reconstruction attributed to the inherent effectiveness of the ABPM model, both in the near- and far-fields.

\subsection{Real Data Results}\label{Sec:real}

Our optical system is illustrated in Fig.~\ref{fig:1} (b) and more details can be found in SM. The first experiment is designed to test the reconstructed quality under the ASM threshold, with a specific propagation distance of 15 to 23 cm. Due to the poor performance of the GS algorithm, its results are omitted. Real data results are shown in Fig.~\ref{fig:experiment1}(a), HoloNet produces blurry edges, while CCNN-CGH over-smooths some areas. In contrast, our method yields sharper edges, finer details, and higher resolution, closely matching the actual values, demonstrating superior reconstruction performance. Fig.~\ref{fig:experiment1}(b) further validates our algorithm's performance across distances from 10 cm to 60 cm for two scenes, showing its effectiveness in both near- and far-fields.

\section{Ablation Study} \label{ablation}
\subsection{The Denoising Block}

\vspace{-3mm}
\begin{table}[h]
  \caption{\small{Ablation study on the denoiser block.}}
  \vspace{-2mm}
  \centering
  \resizebox{.47\textwidth}{!}
{\centering
  \begin{tabular}{cccccc} 
\hline
Denoiser Type  &Resolution& Params&FLOPs(G) & Avg.PSNR(dB) &Avg.SSIM\\ 
\hline
    w/o&1920$\times$1080& -&-&  28.56&0.79\\
  Complex TV  &1920$\times$1080& -&-& 29.54&0.85\\
  Complex CNN  &1920$\times$1080& 22,368&4.46& 32.12&0.90\\
 CVTF &256$\times$256& 114,150& 1.12& 29.09&0.88\\
    CDAT&1920$\times$1080& 19,430&2.63&  \textbf{36.45}&\textbf{0.95}\\
\hline
\end{tabular}}
  \label{tab:denoiser}
\end{table}
   \vspace{-1mm}
   
To validate the effectiveness of our algorithms, we conduct an ablation study on the denoising module, comparing our method with direct removal, model-based complex total variation (TV) algorithm~\cite{gao2023iterative}, learning-based complex CNN algorithm and the existing Complex-Valued Transformer (denoted as ${\rm CVTF}$)~\cite{yang2020complex}.  Due to high computational cost, the resolution in the ${\rm CVTF}$ comparison is limited to $256\times256$. As shown in Tab.~\ref{tab:denoiser}, our algorithm achieves an average PSNR improvement of 4.33-7.89 dB and SSIM enhancement of 0.05-0.16, outperforming other methods and demonstrating that the ${\rm CDAT}$ in our denoising module enables high-quality reconstruction results.

\vspace{-3mm}
\begin{table}[h]
  \caption{\small{Ablation study on the embedding channels.}}
  \vspace{-2mm}
  \centering
  \resizebox{.4\textwidth}{!}
{\centering
  \begin{tabular}{cccc} 
\hline
Embedding Channels & Params(M) & Flops(G)& Avg.PSNR(dB)\\ 
\hline
    96& 87,473&26.8& 36.87\\
 32& 19,430& 2.6&36.45\\
\hline
\end{tabular}}
  \label{tab:embed}
\end{table}
   \vspace{-3mm}

\subsection{Embedding Channel}

In CGH applications, maintaining a lightweight design is essential. As discussed in Sec.~\ref{csda}, the number of embedding channels $C$ is closely related to the computational complexity. As shown in Tab.~\ref{tab:embed}, when the embedding dimension is set to 32, the reconstruction achieves excellent results while keeping the computational cost relatively low.

\section{Conclusion}

To achieve high-quality CGH display, previous CNN-based methods are limited in propagation distance and performance, while fixed global self-attention in complex-valued Transformers is impractical. Towards this end, this paper introduces a deep unfolding network with a Complex-valued Deformable Self Attention mechanism on an Adaptive Bandwidth-Preserving Method. Simulation and optical experiments show that our method significantly outperforms previous approaches in both near- and far-fields.  This work paves the way for high-quality CGH generation with potential applications in optical information storage, tomography, and display technology.

\section*{Acknowledgment}

This work was supported by National Key R$\&$D Program of China (2024YFF0505603), the National Natural Science Foundation of China (grant number 62271414), Zhejiang Provincial Distinguished Young Scientist Foundation (grant number LR23F010001), Zhejiang “Pioneer” and “Leading Goose” R$\&$D Program (grant number 2024SDXHDX0006, 2024C03182), the Key Project of Westlake Institute for Optoelectronics (grant number 2023GD007), the 2023 International Sci-tech Cooperation Projects under the purview of the “Innovation Yongjiang 2035” Key R$\&$D Program (grant number 2024Z126).

\bibliographystyle{IEEEtran}
\bibliography{IEEEabrv,reference}

\end{document}